\documentclass[journal]{IEEEtran}

\usepackage[utf8]{inputenc}
\usepackage{graphicx}
\usepackage{amsmath}
\usepackage{amsthm}
\usepackage{amssymb}
\usepackage{hyperref}
\usepackage{subfig}
\usepackage{multirow}
\usepackage[section]{placeins}

\usepackage{tabularx,booktabs}
\newcolumntype{C}{>{\centering\arraybackslash}X} % centered version of "X" type
\usepackage{lipsum}
\usepackage[table]{xcolor}

\renewcommand\thesection{\arabic{section}}

\begin{document}
\title{Vehicle Driving Assistant}
\author{Akanksha~Dwivedi,
        Anoop~Toffy,
        Athul~Suresh,
        and~Tarini~Chandrashekhar,~\IEEEmembership{International Institute of Information Technology, Bangalore}}% <-this % stops a space

\markboth{Machine Perception Project Report, Group 7, \today}%
{Shell \MakeLowercase{\textit{et al.}}: Bare Demo of IEEEtran.cls for IEEE Journals}

\IEEEspecialpapernotice{(Draft Paper)}

\maketitle

\begin{abstract}
Autonomous vehicles has been a common term in our day to day life with
car manufacturers like Tesla shipping cars that are SAE Level 3. While
these vehicles include a slew of features such as parking assistance and cruise
control, they’ve mostly been tailored to foreign roads. Potholes, and the
abundance of them, is something that is unique to our Indian roads. We
believe that successful detection of potholes from visual images can be applied
in a variety of scenarios. Moreover, the sheer variety in the color, shape and
size of potholes makes this problem an apt candidate to be solved using
modern machine learning and image processing techniques.
\end{abstract}


\IEEEpeerreviewmaketitle

\section{Introduction}
The project aims to provide a comprehensive set of assistance features to aid the driver (or autonomous vehicle) to drive safely. This could includes a number of indicators about the environment, the major cue being visualisation and detection of potholes in the road ahead. 

"A pothole \cite{pothole} is a structural failure in a road surface, caused by failure primarily in asphalt pavement due to the presence of water in the underlying soil structure and the presence of traffic passing over the affected area". An example of a road with pothole is shown in Figure 1.

\begin{figure}[!h]
\begin{center}
\includegraphics[scale=0.065]{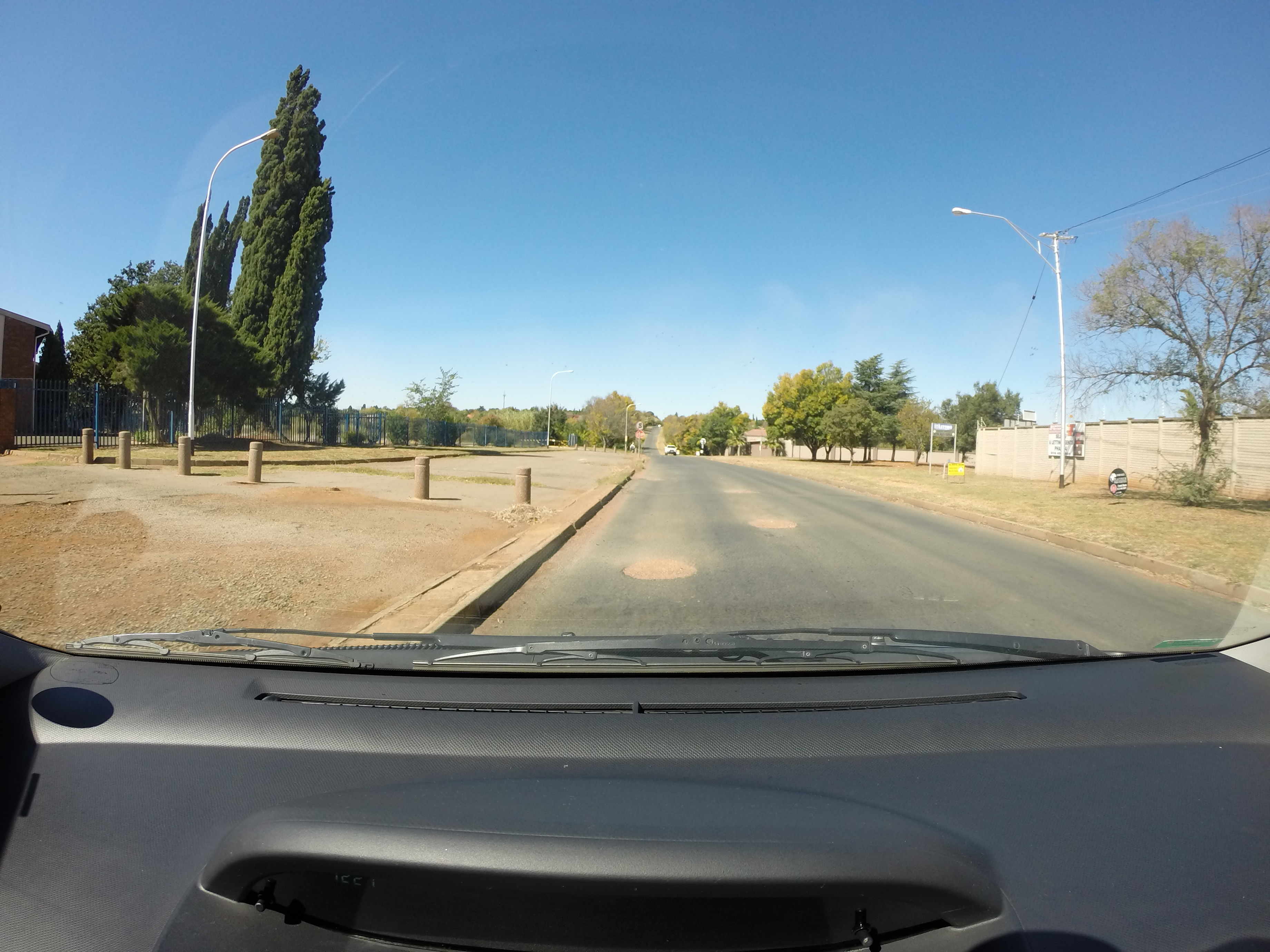}
\end{center}
\caption{Pothole example}
\end{figure}

The remainder of the paper is structured as follows. We present a brief history of the related work done so far in the field of pothole detection and visualisation in Section 2. Section 3 discusses about the experimental set up we used in this work.In the subsequent section we describes the methodology that we follow in the following section 4. It includes road extraction part and blob detection part. Then the result that we obtained are given briefly in section 5. Finally section 6 concludes the paper. Section 7 gives a light to the future enhancements that could be done.

\section{Related Work}
In Nienaber et al (2015) \cite{paperone}, a system using basic image processing techniques in a constrained environment without relying on any machine learning techniques is used for pothole detection. It presents a good preliminary method for detecting potholes using a single camera within an range of 2 - 20m from a vehicle moving at a speed of not more than 60km/hr. The method separates a rectangular area of interest just above the hood of the vehicle which contains road surface, assuming that driver maintains a safe distance from the front vehicle. The rectangular area of interest is separated by connecting the various farthest region of interest using convex hull algorithm.

\vspace*{.5cm}

The work presented by Ajit Danti et al (2012) \cite{papertwo}, presents a comprehensive approach to address the acute problems of Indian roads such as faded lanes, irregular potholes, improper and invisible road signs. Instead of using image processing techniques for pothole detection as done by Nienaber et al (2015), Ajith Danti et al (2012) uses K-Means clustering based algorithm to detect potholes. By addressing the acute problems above mentioned in the paper it makes automated driving safer and easier in Indian roads. 

\section{Proposed Work}
In this paper we use both image processing techniques as well as machine learning to study the presence and occurrence of potholes. High resolution image captured from the hood of a slow moving vehicle is used for analysis and experimentation. This paper propose a way to visualize the potholes as well as identify whether or not the road has a pothole in it. The visualisation approach is similar to how humans perceive the pothole images in the brain. The pothole identification approach will help to signal the driver of a vehicle to take preventive actions upon pothole infected lanes. 

\vspace{0.5cm}

Foliage, incoming vehicles and other background objects can interfere with the detection of potholes. In all the methods described below the area of the image containing the road is isolated and extracted. Subsequent processing is then done on this extracted area to detect the presence of potholes. 

\vspace{0.5cm}

The subsequent subsection describes the approaches we took for visualising the potholes. We used contour detection after applying morphological transformation as a method I and then we used contour detection after extracting the road area by using convex hull algorithm in method II. The methods are described in detail below.

\subsection{Using morphological transformation}

Morphological transformations are some simple operations based on the image shape. It is normally performed on binary images. Two basic morphological operators are Erosion and Dilation. Using which the image is analysed to visualise the pothole present in the region.

\subsection{Using contours and convex hull}

For visualising the pothole we first extract a region of the interest from the given image which will contain the road area. This is done by first selecting the largest contour on the images ROI and then drawing convex hull using convex hull algorithm. Since extracting the road area will help to narrow down the area of interest. This way we will be able to prevent the irrelevant informations from the analysis like foliage, vehicle hood, other vehicles on the road etc. We then apply image processing techniques like contour detection, blob detection and edge detection to draw a bounding box over the area of the road which contains the potholes.

\subsection{Using Machine Learning Methods}
For machine learning technique we used annotated dataset \cite{dataset} for training various classifiers and then try various machine learning algorithms to figure out the presence of a pothole in the frame given.

\section{Methodology}

In this section we briefly familiarize the methodology that we used for the visualisation and detection of potholes in detail. The subsequent section first discusses about the image processing techniques that are helpful for pothole visualisation followed by feature extraction and machine learning learning techniques  used in the study.

\subsection{Pothole Visualisation Techniques}

\subsubsection{Morphological Transformation Method}
The image is smoothed using Gaussian blur and then thresholding is applied. Then opening and closing morphological transformation are applied to clear off the noise in the image and then road area is extracted from the output image. From the extracted road area we applied edge detection methods to find potholes present in the image. This technique works on the Google Images dataset with partial noise being detected but detected too much noise in our dataset along with a fraction of potholes.

\begin{figure}[!htb]
\begin{center}
\includegraphics[scale=0.65]{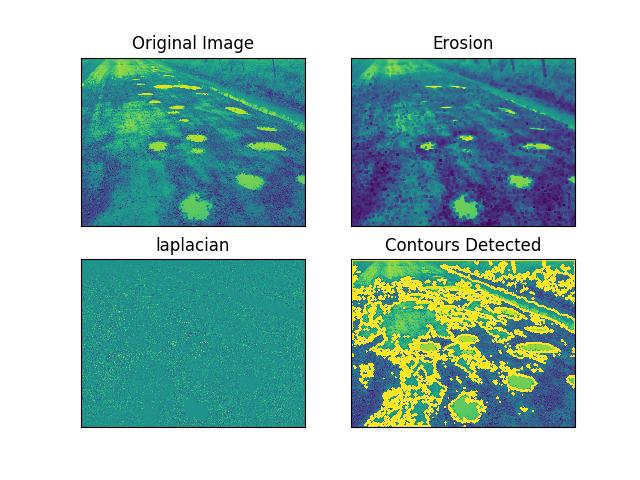}
\end{center}
\caption{Results of Morphological Transformation}
\end{figure}

\newpage

\subsubsection{Visualisation from isolated roads}

\noindent The figure 3 show the image which is collected from the Digital camera. The image is then converted to its corresponding HSV channel because HSV channels captures the most variance in the image which aids is subsequent processing.  

\vspace{0.5cm}

\begin{figure}[!htb]
\begin{center}
\includegraphics[scale=1]{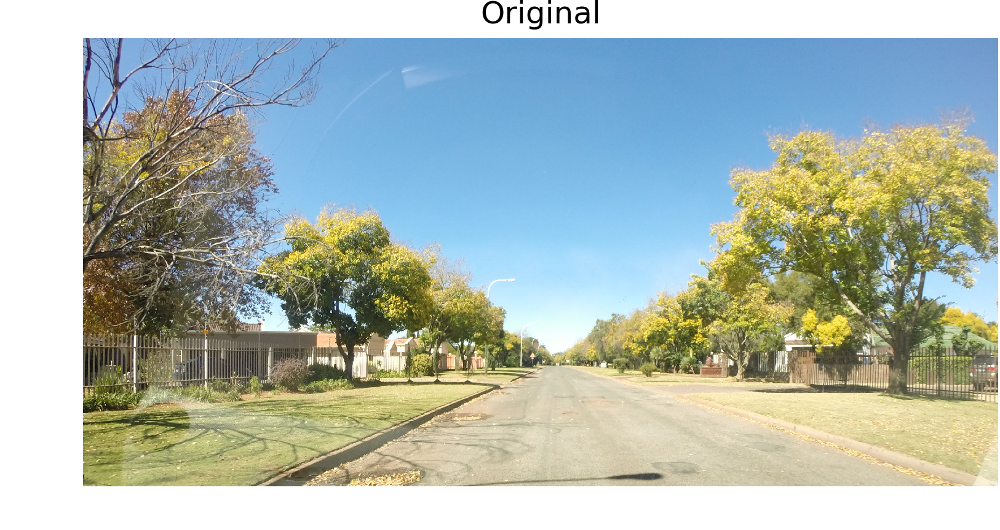}
\end{center}
\caption{Pothole Image}
\end{figure}

\begin{figure}[!htb]
\begin{center}
\includegraphics[scale=1]{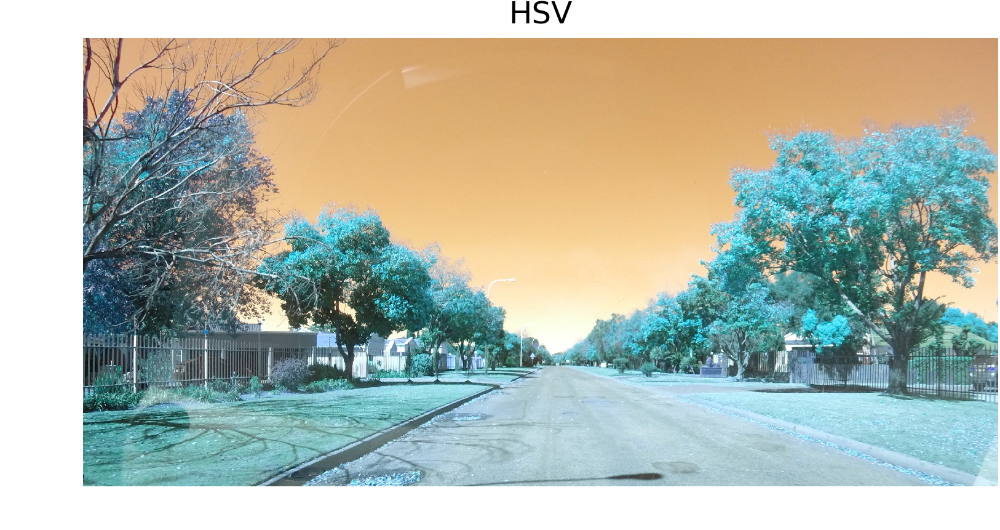}
\end{center}
\caption{Converting to HSV}
\end{figure}

\vspace{0.5cm}

\noindent From the image we select a region of interest (ROI) which contains the road area. Since we are using a fixed camera mounted on the window of a moving vehicle we assume that the road area tend to appear on a fixed region above the hood of the vehicle. Hence we select a Region of Interest from the (ROI) image as show in figure 5.

\begin{figure}[!htb]
\begin{center}
\includegraphics[scale=1]{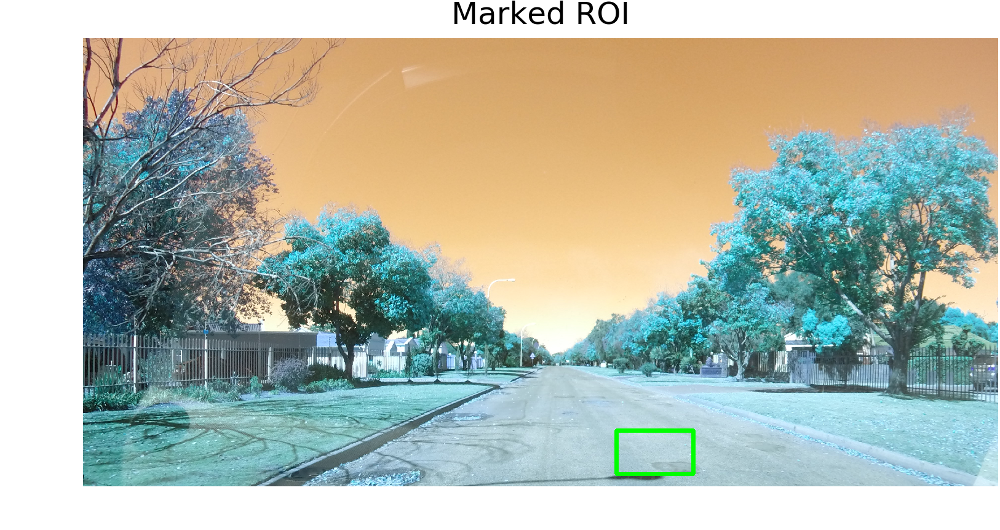}
\end{center}
\caption{Selecting the ROI}
\end{figure}

\vspace{0.5cm}

% add modification here
\noindent Figure 6 shows the enlarged ROI of the image. The mean and standard deviation per colour channel of this region of interest is calculated. In order to address the issue of road colour variation within the image, the road colour is modelled separately per channel as lying within three standard deviations of the mean of each channel. These thresholds were used to binarize the image, and the mask obtained is show in Figure 7. From the figure, it is clear that the average colour model does not match the entire road area as there are many areas of the road that were not included after the thresholding operation.
Contour detection algorithm is then run on the extracted mask. It is assumed that the largest contour in the image will include the road and therefore only this contour is sent to the convex hull algorithm.
The largest contour is received by the convex hull algorithm and the output of the algorithm is indicated in Figure 8.  Using a mask made of the computed convex hull, the road is extracted from the input image as shown in Figure 10.

\begin{figure}[!htb]
\begin{center}
\includegraphics[scale=0.65]{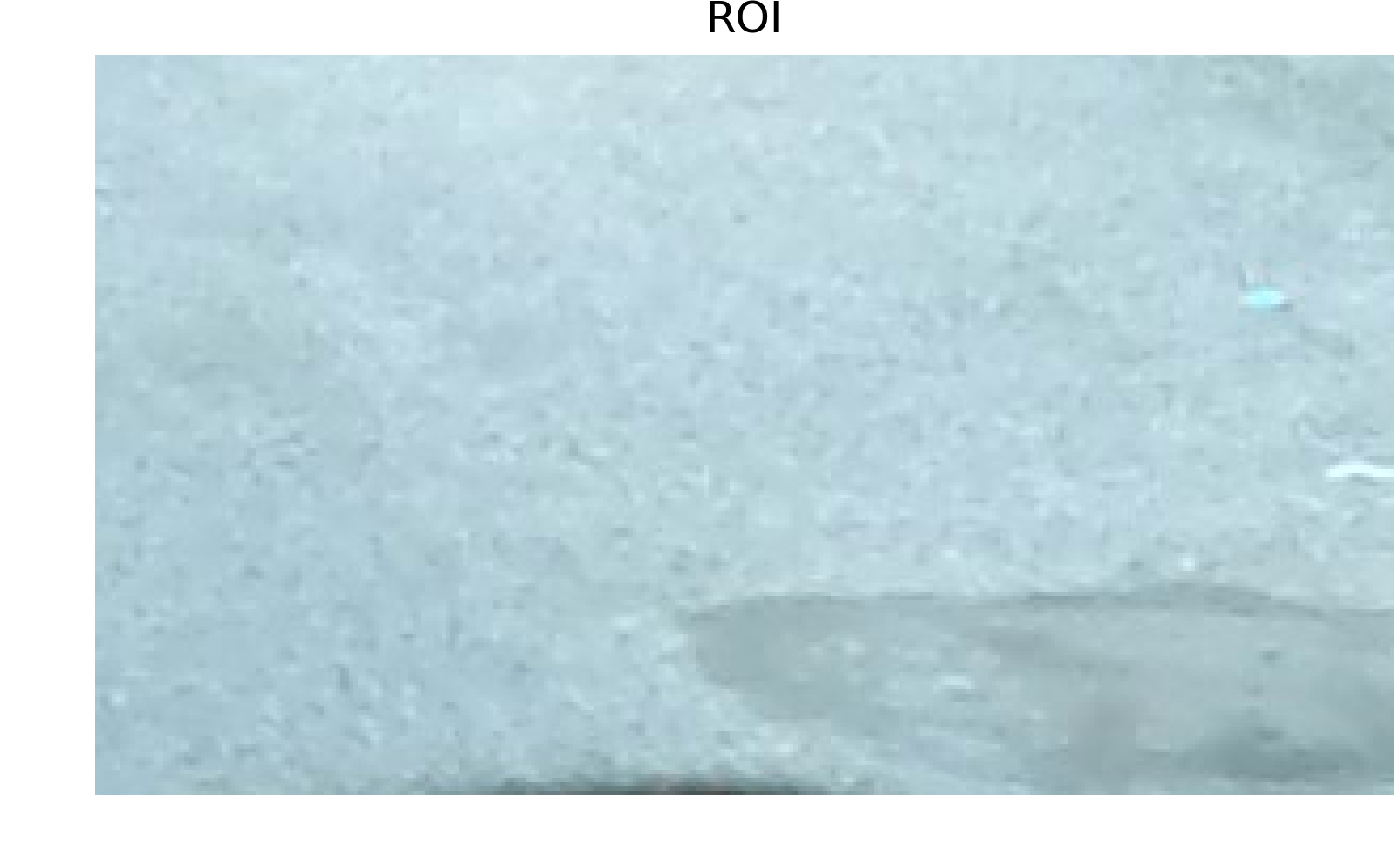}
\end{center}
\caption{Region of Interest}
\end{figure}

\begin{figure}[!htb]
\begin{center}
\includegraphics[scale=0.65]{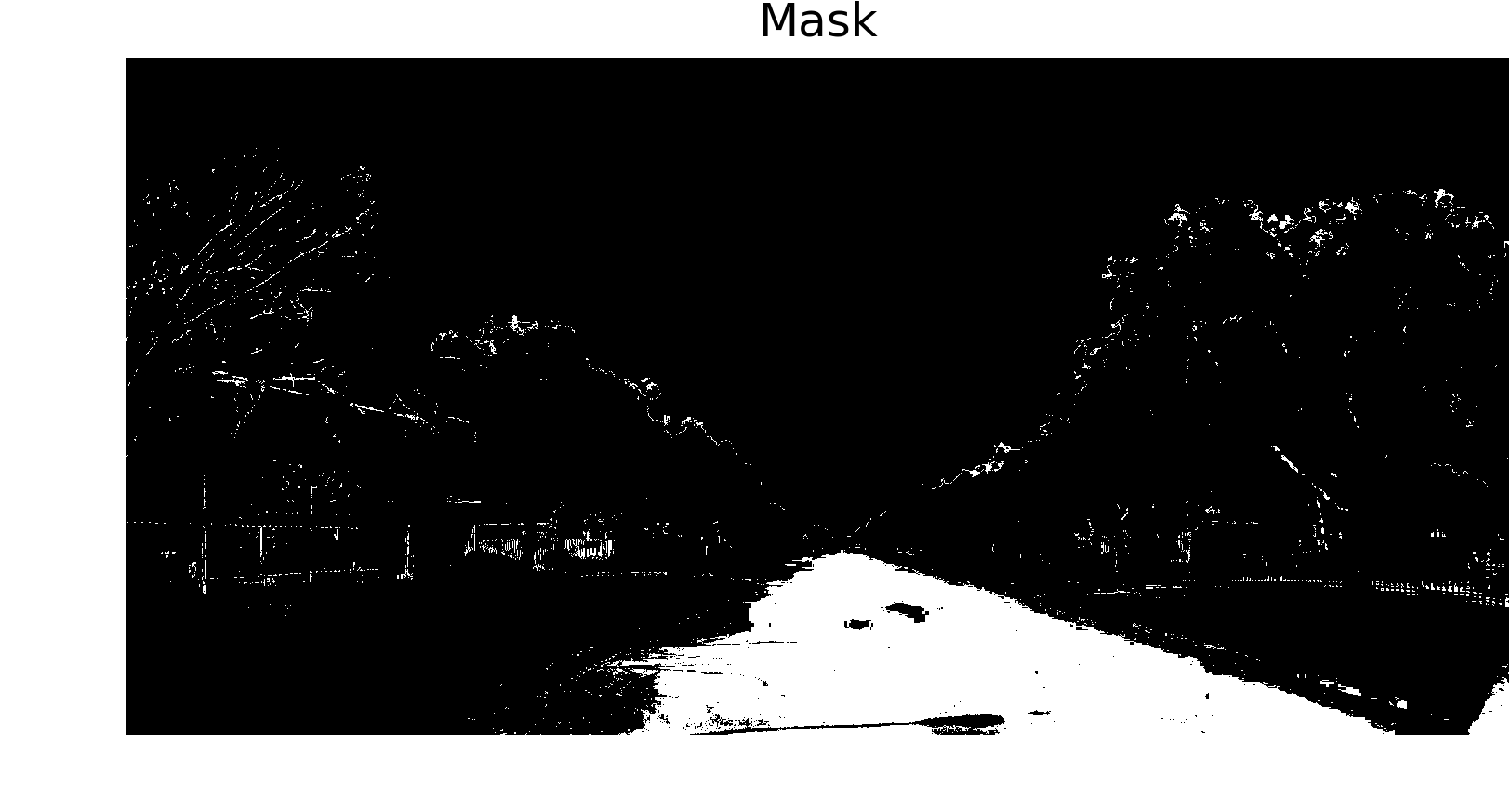}
\end{center}
\caption{Mask}
\end{figure}

\vspace{0.5cm}

\noindent The Figure 8 below indicates the convex hull of the largest contours selected from the threshold images. This largest contour is used a mask to extract the road area as shown in the Figure 10.

\begin{figure}[!htb]
\begin{center}
\includegraphics[scale=0.65]{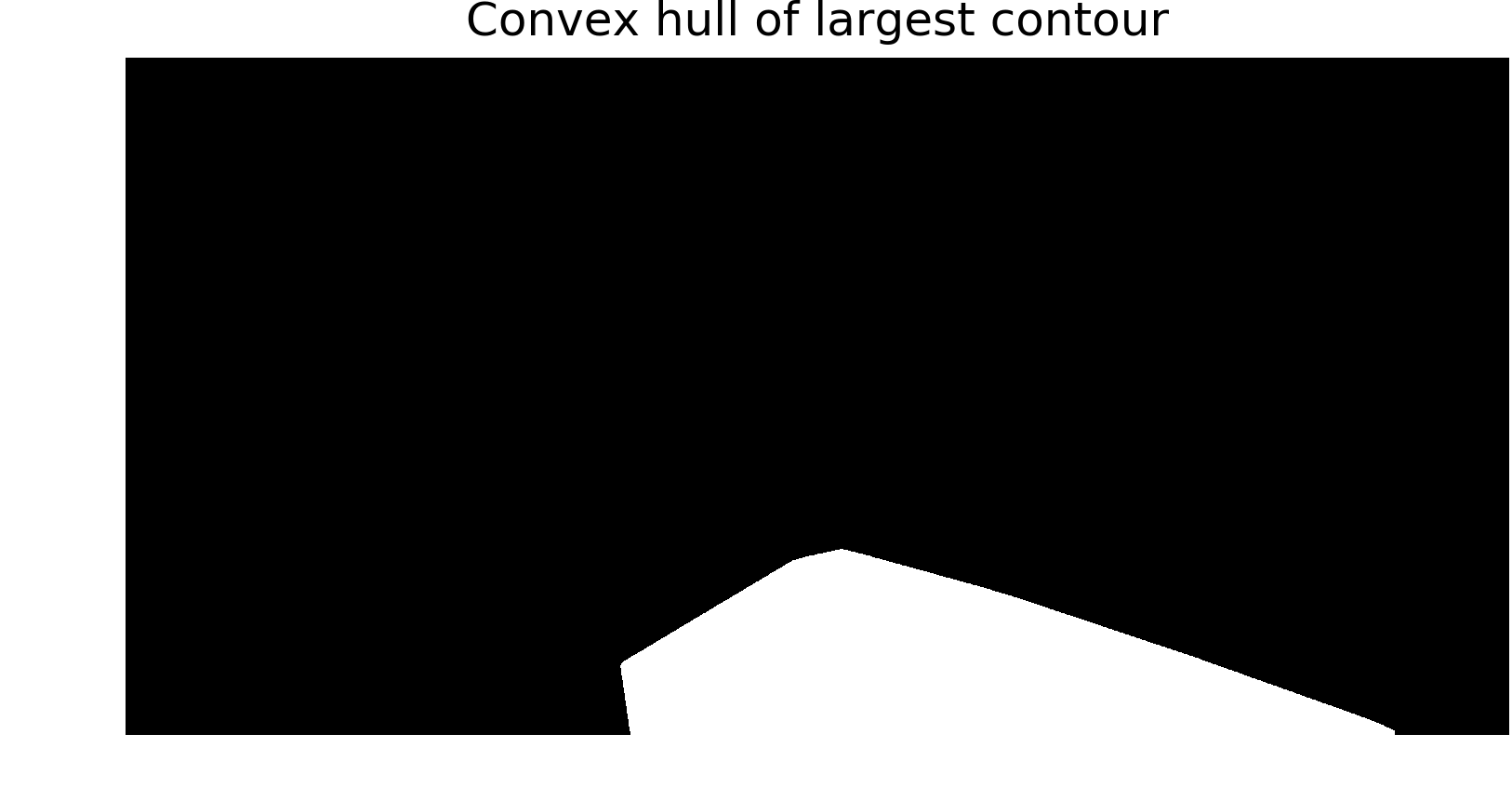}
\end{center}
\caption{Convex hull of largest contour}
\end{figure}

\begin{figure}[!htb]
\begin{center}
\includegraphics[scale=1]{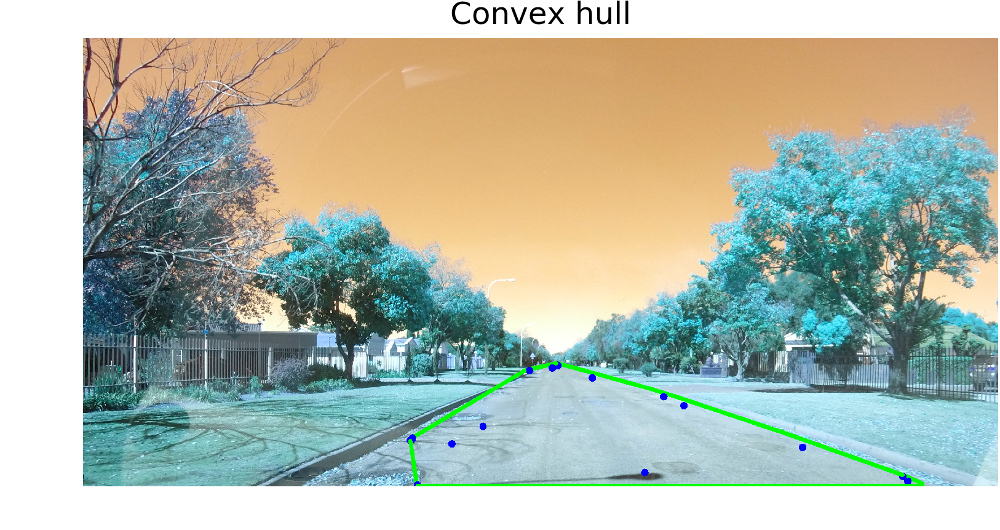}
\end{center}
\caption{Convex hull on Original Image}
\end{figure}
\newpage

\begin{figure}[!htb]
\begin{center}
\includegraphics[scale=0.65]{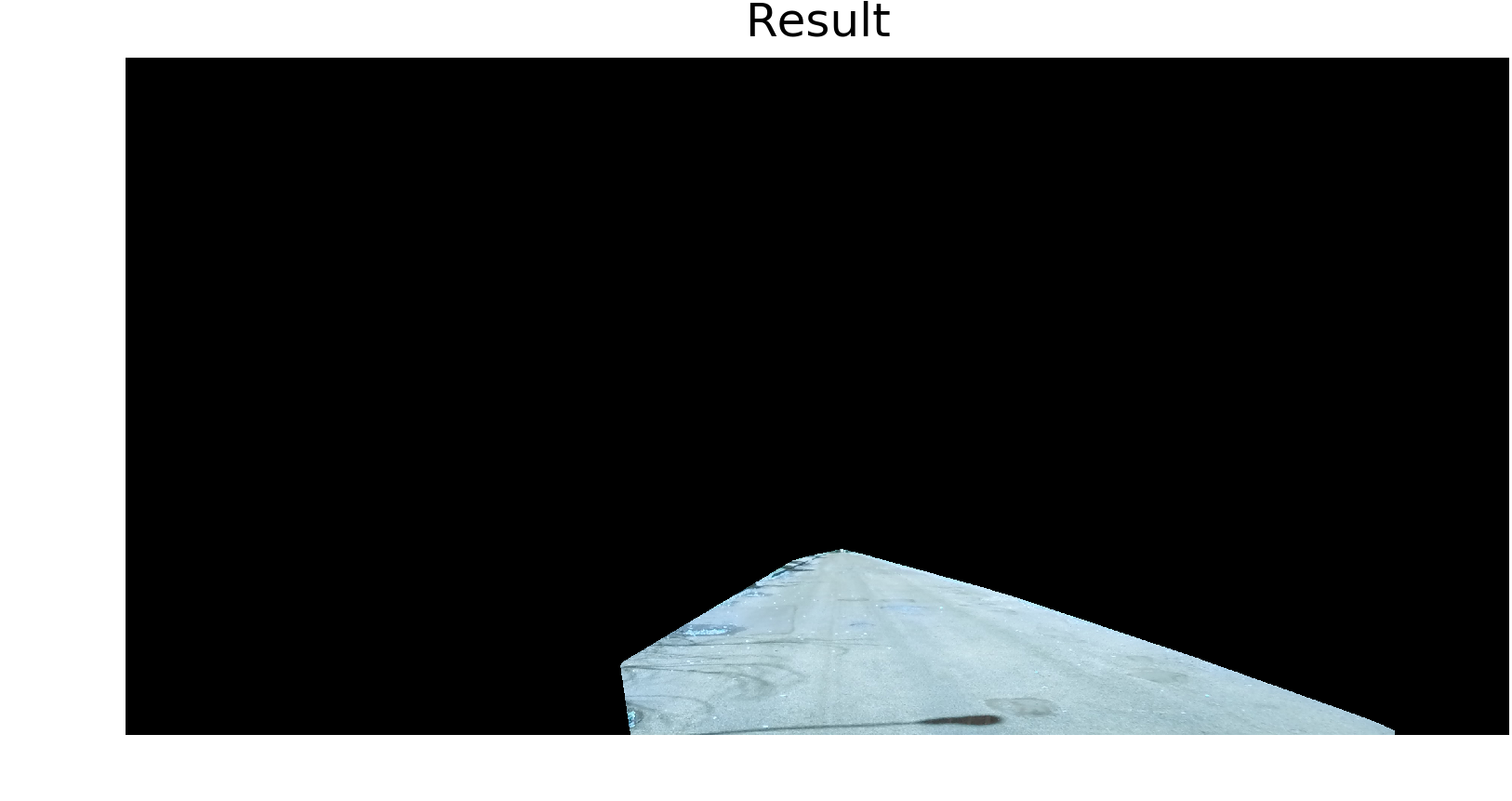}
\end{center}
\caption{Extracted Road}
\end{figure}

\vspace*{.5cm}

\subsubsection*{Blob Detection Method}

The extracted road area is then analysed further for pothole visualisation. In computer vision, blob detection methods are aimed at detecting regions in a digital image that differ in properties, such as brightness or color, compared to surrounding regions. Since we assume that pothole appear in the road in different texture than that of the road area mainly in dark shadowed regions some time it appear as though it is covered in mud and water. Since there is a differentiating factor that make pothole identifiable in the road we used blob detection algorithms to highlight the potholes.

\vspace*{.5cm}

\subsubsection*{Contour dilation Method}
The extracted road area could also be analysed using contour detection or edge detection (Canny edge detection). The detected edges could be dilated for highlighting the presence of potholes. But the results obtained for the potholes were sub\-optimal. Since potholes appear in varied shape and color better discrimination from other object on the road could not be achieved. 

\subsection{Pothole detection using machine learning techniques}

In this section we describe in details the various machine learning techniques that we used for detecting the presence of the pothole in the roads. As described in the previous section, the ROI containing the road is extracted using the previously mentioned Convex hull of contour method. This ROI is then subsequently processed by the models to detect the presence of potholes.

\subsubsection{Feature Modelling}

The feature extraction is done in two ways 
\begin{itemize}
\item Downscaled image as features \\
The image is downscaled by a factor of two and then flattened to be used as a feature vector.
\item Color histogram as a feature \\
A color histogram of an image represents the distribution of the composition of colors in the image. It shows different types of colors appeared and the number of pixels in each type of the colors appeared.

\vspace*{0.5cm}

"The histogram provides a compact summarization of the distribution of data in an image. The color histogram of an image is relatively invariant with translation and rotation about the viewing axis, and varies only slowly with the angle of view. By comparing histograms signatures of two images and matching the color content of one image with the other, the color histogram is particularly well suited for the problem of recognizing an object of unknown position and rotation within a scene" \cite{hist}
% add here.
\end{itemize}

\subsubsection{Machine Learning Algorithms}

The following list of machine learning techniques were used in the study. They are trained using the above mentioned feature and the results are analysed.

\begin{itemize}
\item KNN \cite{knn}
\item GaussianNB
\item Logistic Regression \cite{lr}
\item SVM \cite{svm}
\item Decision Tree \cite{dt}
\item Random Forest \cite{rf}
\item AdaBoost using Decision Tree
\end{itemize}

\section{Results}

\subsection{Pothole visualisation using Image processing Techniques}

The image processing techniques gave us suboptimal results. Method I using morphological transformation didn't work well as there are edges in the roads that created extra contours which were wrongly highlighted as potholes. Method II using road isolation followed by contour detection and edge detection worked for some scenarios but did not work well for most of the cases. Hence pothole visualisation is found difficult to achieve with the current technology as we were not able to find discriminating feature for potholes from our analysis

\subsection{Pothole detection using Machine Learning Techniques}

The classifier are trained using the feature and the results obtained are shared in the below tables.
The table below show the result obtained while using the downscaled image as features.
\begin{center}
\begin{table}[h!]
\centering
\begin{tabular}{ |c|c| } 
 \hline
 \rowcolor{gray}
 Methods & Accuracy($\%$)  \\ 
 \hline
 Logistic Regression & 72.73 \\
 \hline
 Multinomial Logistic Regression & 68.93\\
 \hline
 Decision Tree & 86.88 \\ 
 \hline
 AdaBoost using Decision Tree & 80.32  \\ 
 \hline
 GaussianNB & 44.26 \\
 \hline
 KNN & 73.77\\
 \hline
 SVM & 55.73\\
 \hline
 Random Forest & 68.18\\
 \hline
\end{tabular}
\caption{Results for Raw downscaled images pixels as features}
\label{table:1}
\end{table}
\end{center}

\vspace{0.5cm}

The table below show the result obtained while using the color histogram as features.
\begin{center}
\begin{table}[h!]
\centering
\begin{tabular}{ |c|c| } 
 \hline
 \rowcolor{gray}
 Methods & Accuracy($\%$)  \\ 
 \hline
 Logistic Regression & 77.27\\
 \hline
 Decision Tree & 77.04 \\ 
 \hline
 AdaBoost using Decision Tree & 83.60  \\ 
 \hline
 GaussianNB & 75.40 \\
 \hline
 KNN & 78.68\\
 \hline
 SVM & 55.73\\
 \hline
 Random Forest & 95.45\\
 \hline
\end{tabular}
\caption{Results for color histogram as features}
\label{table:1}
\end{table}
\end{center}

As we can infer from the tables that the classification using downscaled images worked well when decision trees and AdaBoost using decision tree are used. Meanwhile Gaussian Naive Bayes performed the worst using downscaled images.

\vspace{0.5cm}

The color histogram as a feature worked relatively better when compared to the downscaled images. We were pretty surprised to see the performance of color histogram using Random Forest which gives an accuracy of 
95$\%$. SVM didn't give good accuracy in both color histogram as well as downscaled images.

\section{Conclusion}
We presented a mix of methods, incorporating both image processing and machine learning, to detect the presence of potholes in an image. Given an image, the road was isolated and extracted with best results using contours and finding the convex hull of the largest contour. Through successive testing, it was found that using machine learning models yielded better results than conventional image processing methods as far as detection of potholes were concerned. The models were trained using features extracted from the isolated road segment. Comparison between the different methods of feature extraction and the performance of different models on these features have also been exhibited.

\section{Future Work $\&$ Improvements}
The present work yields sub\-optimal results in scenes features heavy shadows. Better discriminating features could be extracted which helps to discern potholes even in environments where the lighting is uneven and interspersed with shadows. Methods could be developed to accommodate the presence of incoming traffic and road markings.

\section*{Acknowledgment}
We would like to thank Prof. Dinesh Babu Jayagopi for guiding us throughout this venture, Prof. MJ (Thinus) Booysen, Associate Professor at the Electrical $\&$ Electronic Engineering Department at Stellenbosch University for prividing us with the pothole datasets \cite{dataset}.

\vspace{0.5cm}

\ifCLASSOPTIONcaptionsoff
  \newpage
\fi

\appendices
\section{Data and Code}
\begin{itemize}
\item Data 
\begin{itemize}
\item Data Set I - \url{https://www.youtube.com/watch?v=hmeBmdZzlLU}
\item Data Set II - \url{https://www.youtube.com/playlist?list=PLzf2LOyk6iN3nNyvu1RI0Np-i8kT1IGE0}
\item Data Set III \cite{dataset}
\end{itemize}

\item Code - \url{https://github.com/crunchbang/MP_Project/blob/master/Code/}
\item Road Extraction Video - \url{https://www.youtube.com/watch?v=4vUqzDnZsV8}
\end{itemize}

%\section{}

\newpage

\begin{IEEEbiography}[{\includegraphics[width=1in,height=1.25in,clip,keepaspectratio]{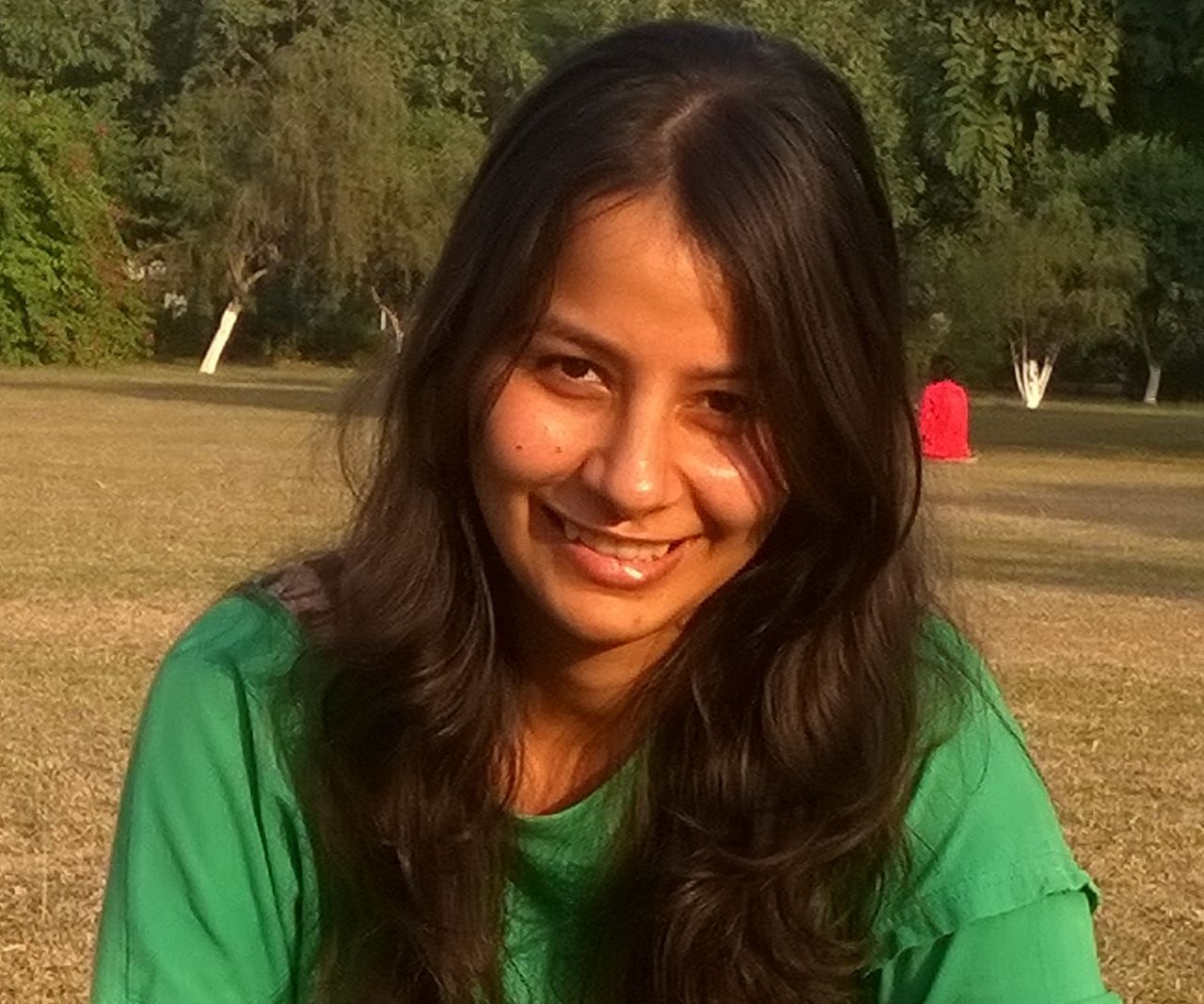}}]{Akanksha Dwivedi}
(MT2016006) is currenlty pursing Master of Technology from International Institute of Information Technology, Bangalore. 
\end{IEEEbiography}

\begin{IEEEbiography}[{\includegraphics[width=1in,height=1.25in,clip,keepaspectratio]{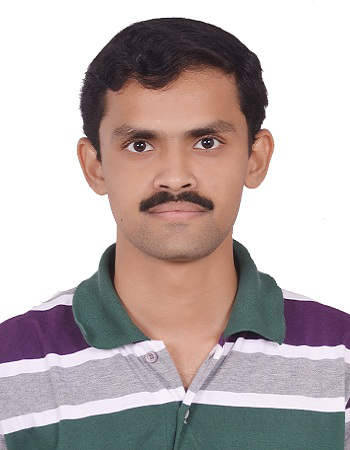}}]{Anoop Toffy}
(MT2016016) is currenlty pursing Master of Technology from International Institute of Information Technology, Bangalore. 
\end{IEEEbiography}

%\newpage

\begin{IEEEbiography}[{\includegraphics[width=1in,height=1.25in,clip,keepaspectratio]{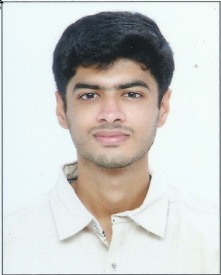}}]{Athul Suresh}
(MT2016030) is currenlty pursing Master of Technology from International Institute of Information Technology, Bangalore. 
\end{IEEEbiography}

\begin{IEEEbiography}[{\includegraphics[width=1in,height=1.25in,clip,keepaspectratio]{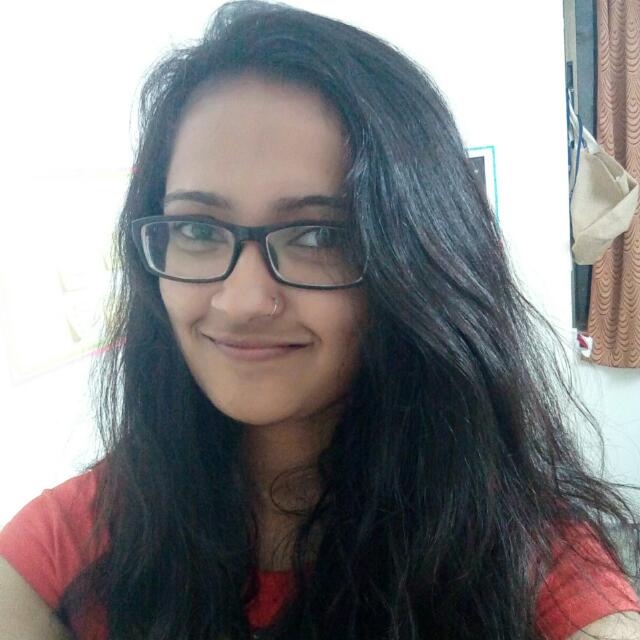}}]{Tarini Chandrashekhar}
(MT2016144) is currenlty pursing Master of Technology from International Institute of Information Technology, Bangalore. 

\end{IEEEbiography}


\begin{thebibliography}{1}

\bibitem{paperone} 
S. Nienaber, M.J. Booysen, R.S. Kroon
\textit{Detecting potholes using simple image processing techniques and Real-world Footage, 2015}. 
SATC, July 2015, Pretoria, South Africa
\url{http://scholar.sun.ac.za/handle/10019.1/97191}
 
\bibitem{papertwo} 
Ajit Danti, Jyoti Y. Kulkarni, and P. S. Hiremath, Member, IACSIT
\textit{An Image Processing Approach to Detect Lanes, Pot Holes and Recognize Road Signs in Indian Roads, December 2012}
\url{http://www.ijmo.org/papers/204-S3015.pdf}

\bibitem{paperthree}
S. Nienaber, R.S. Kroon, M.J. Booysen  
\textit{“A Comparison of Low-Cost Monocular Vision Techniques for Pothole Distance Estimation”}
IEEE CIVTS, December 2015, Cape Town, South Africa.
 
\bibitem{dataset}
The annotated image dataset used in the pothole detection is freely available at
\url{https://goo.gl/3QyeMs}

\bibitem{pothole}
\url{https://en.wikipedia.org/wiki/Pothole}

\bibitem{lr}
\url{https://en.wikipedia.org/wiki/Logistic_regression}

\bibitem{dt}
\url{https://en.wikipedia.org/wiki/Decision_tree}

\bibitem{knn}
\url{https://en.wikipedia.org/wiki/K-nearest_neighbors_algorithm}

\bibitem{rf}
\url{https://en.wikipedia.org/wiki/Random_forest}

\bibitem{svm}
\url{https://en.wikipedia.org/wiki/Support_vector_machine}

\bibitem{dataset1}
Dataset with negative and positive examples separated
\url{https://drive.google.com/drive/folders/0B7LHCitTUdEYZFEwNWo4V2RldjQ?usp=sharing}

\bibitem{opencv}
Open CV is used for Image processing and Machine Learning
\url{http://docs.opencv.org/}


\bibitem{hist}
Color Histogram
\url{https://en.wikipedia.org/wiki/Color_histogram}

\end{thebibliography}
\end{document}